43# When is Early Classification of Time Series Meaningful?

4Renjie Wu, Audrey Der, and Eamonn J. Keogh

44
**Abstract**—Since its introduction two decades ago, there has been increasing interest in the problem of *early classification of time series*. This problem generalizes classic time series classification to ask if we can classify a time series subsequence with sufficient accuracy and confidence after seeing only some prefix of a target pattern. The idea is that the earlier classification would allow us to take immediate action, in a domain in which some practical interventions are possible. For example, that intervention might be sounding an alarm or applying the brakes in an automobile. In this work, we make a surprising claim. In spite of the fact that there are dozens of papers on early classification of time series, it is not clear that any of them could ever work in a real-world setting. The problem is not with the algorithms per se but with the vague and underspecified problem description. Essentially all algorithms make implicit and unwarranted assumptions about the problem that will ensure that they will be plagued by false positives and false negatives even if their results suggested that they could obtain near-perfect results. We will explain our findings with novel insights and experiments and offer recommendations to the community.

✦

## 1 INTRODUCTION

SINCE its introduction two decades ago, there has been increasing interest in the problem of *early classification of time series* (ETSC). The problem is expressed differently by different researchers, but it generally reduced to asking if we can classify a time series subsequence with sufficient accuracy and confidence after seeing only some prefix of a target pattern. Using text as an analogy for time series, if someone typed `albuquer...`, we could be very confident that they planned to type the name of the most populous city in New Mexico.

The key claim is that classification without waiting for the entire pattern to appear would allow us to take immediate action in a domain in which some interventions are possible. For example, that intervention might be pretightening the seatbelts in an automobile that the classifier predicts may be about to crash.

While the idea of ETSC is interesting and socially noble, in this work, we make a somewhat surprising claim. In spite of the fact that there are many research efforts on ETSC, it is not clear that any of them could ever work in a real-world setting. The problem is not with the algorithms per se but with the vague and underspecified problem description. Most of the issues stem from a mismatch between the data format used to train and test ETSC models and the data format that must be used in the real world. Most ETSC papers consider only data in the UCR format, which as shown in Fig. 1, assuming that all exemplars are of the same length and at least approximately aligned in time [1].

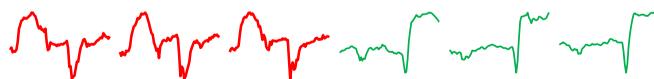

Fig. 1. Samples of data in the UCR format. Note that exemplars are all of the same length and carefully aligned. The exemplars are utterances of the words cat and dog, spoken by a female in Standard American English, represented in MFCC Coefficient 2.

Given data formatted in this way, the ETSC community has produced dozens of models that can predict the class of an incoming subsequence, after only seeing a fraction of the data [2], [3], [4], [5], [6], [7], [8], [9], [10], [11], [12], [13], [14], [15], [16], [17], [18], [19], [20]. This sounds impressive, but as shown in Fig. 2, consider what would happen when we test on the utterance "*It was said that Cathy's dogmatic catechism dogmatized catholic doggery*".

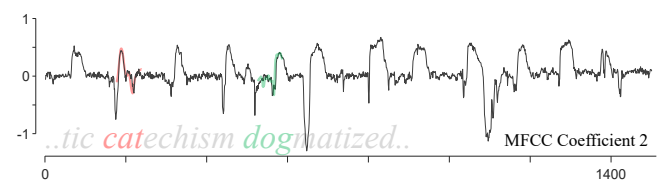

Fig. 2. A snippet of the phrase "*It was said that Cathy's dogmatic catechism dogmatized catholic doggery*". This short sentence will allow any ETSC method to make confident and early predictions, all of which will later have to be recanted.

This sentence will produce six false positives: three in each class. Note that we cannot brush the problem aside by saying that we can simply recant the classifications *after* we see the rest of the longer word. The whole point of ETSC is to take some actions. The action might be "just" sounding an alarm, but even just false alarm fatigue is known to have a huge cost [21]. If 99.9% of all

___________

- R. Wu is with the Department of Computer Science and Engineering, University of California Riverside, Riverside, CA 92521. E-mail: rwu034@ucr.edu.
- A. Der is with the Department of Computer Science and Engineering, University of California Riverside, Riverside, CA 92521. E-mail: ader003@ucr.edu.
- E.J. Keogh is with the Department of Computer Science and Engineering, University of California Riverside, Riverside, CA 92521. E-mail: eamonn@cs.ucr.edu.

alarms are false positives, it seems inconceivable that the system would be used.

It is also important to recall that by the explicit definition of the ETSC problem, the action must be *immediate*. If we wait "to make sure", then in no sense are we doing *early* classification — we are just doing classification.

This issue of false positives would be damning even if we had no false negatives. However, as we will show in Section 4, most ETSC methods have a misunderstanding about the normalization of the data that will condemn them to produce many false negatives.

We call the "cat" vs "cat*alog*" problem the *prefix* problem. We will later show two other issues, the *inclusion* and *homophone* problems that offer even greater stumbling blocks to any ETSC models.

The absolute weakest interpretation of our findings is that the ETSC community has failed to communicate or appreciate the many assumptions that must be true for their models to be useful in the real world. However, we will argue a stronger interpretation. The ETSC problem is underspecified to the point of being meaningless, and the entire area needs to be "rebooted" with greater rigor.

## 2 BACKGROUND

### 2.1 How ETSC Algorithms Work

The idea of early classification in time series seems to have originated in an obscure paper in 2001 [22], however the problem framework that is most commonly understood appears in a sequence of papers by Xing *et al.* [3], [4]. These works define the challenge as finding the best compromise between accuracy of prediction and earliness of prediction in the face of incrementally arriving data. This can be framed in several ways, and different papers use slightly different terminology. However, Fig. 3 shows the two most common interpretations of this idea.

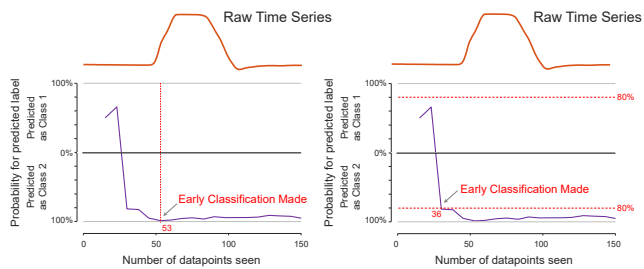

Fig. 3. *(left)* The TEASER model [2] correctly predicts the class of an exemplar from GunPoint after seeing only 53 data points. *(right)* Other models predict only when a user-specified confidence threshold is met.

In Fig. 3 *(left)*, we used the method in [2], working on the ETSC community's favorite dataset, GunPoint [1]. As the data arrives, some models predict the probability that we are seeing the prefix of any of the classes we have trained on. At some point, an internal model decides it has seen enough to trigger a classification. Different papers use different internal models, and a handful incorporates some awareness of misclassification costs [12], [19]. In Fig. 3 *(right)*, we see another common framing of the problem. Here the ETSC algorithm simply predicts the probability of being in each class, and if that probability exceeds some user-specified threshold. In this case, the user's threshold of 0.8 allowed classification after seeing only 36 datapoints. In a sense, the two models are equivalent, and the slight distinctions do not concern us here.

### 2.2 Disconnect to the Real World

The motivation for early time series classification is plausible, although to our knowledge there has never been an ETSC algorithm deployed in the real world. As we will see, this seems to be a telling fact. In contrast, while classic time series classification is perhaps an overstudied problem, it is still easy to point to hundreds of commercial and scientific applications that actually use it.

One issue seems to be that there is a disconnect between the models and the claimed uses for them. Consider [20], which motivates ETSC with "*in the early diagnosis of heart disease, abnormal ECG signals may indicate a specific heart disease that needs immediate treatment. If a classification model that can make early diagnosis as soon as early of ECG time series is available, the patient with the heart disease can get early treatment.*" As shown in Fig. 4, the authors of [20] do indeed test on ECGs.

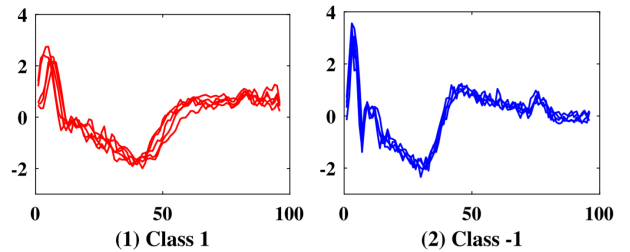

Fig. 4. A screen dump from [20]. The authors test on an ECG dataset from the UCR Archive [1].

They later also correctly note "*If a person has a myocardial infarction, it is usually observed from the ECG that the ST wave is changed and elevated…*". However, let us step back a moment. Yes, it is true that heart disease needs immediate treatment. However, that is typically understood at the scale of "*today*" is better than "*next month*". Maybe the authors meant the case of a patient recovering in an ICU with the plan to page a doctor the moment we see a single myocardial infarction. The full ECG beats in question are about 0.5 seconds long. Suppose, as [20] claims, we could classify the abnormal heartbeats after seeing only 64% of the data. That means that we could alert the doctor 0.18 seconds earlier. This is an inconsequent amount, especially for a warning that comes with a 17% chance of being a false positive [20] (as we will see in Section 4, the claim is in any case spurious, as it makes a normalization assumption that could not be true).

More generally, there does seem to be a disconnect in the literature between the obvious and true motivation that "earlier is better", and any practical actionable application of ETSC. In any case, this discussion may be largely moot because as we will show in the next two sections, no current ETSC algorithm is likely to work in any real-world settings due to three types of confounding issues that the community has not noticed.

## 3 ETSC IS MUCH HARDER THAN IT APPEARS

In Fig. 2, we hinted at a problem caused by assuming that data forced into the UCR format represents a real-world problem. As damning as this single issue is, we will now demonstrate that it is only one of the three related issues that cast doubt not only on the solutions proposed for ETSC but on the very problem definition itself.

### 3.1 The Prefix Issue

**The prefix problem** is the assumption that the pattern to be early classified is not a prefix of a longer innocuous pattern.

Imagine that we have two classes which are the MFCC representation of the spoken words, *cat* and *dog*. Again, under the UCR formatting assumption, this would be an ideal ETSC problem. However, as illustrated in Fig. 2, we need to consider what will happen when we deploy in a streaming environment. Suppose we encountered the perfectly valid sentence "*…all oxen excel at persistence, strength and doggedness. The use of cattle for draft work in…*" [23]. We would get two early classifications, which must then later be recanted.

The reader might imagine that while we may produce an early classification for "*ca…*", we can later retract that prediction when we subsequently see "*…ttle*". But recall that the whole point of early classification is to give actionable early warning. If it is supposed to be actionable, do we take that action or not? If we need to wait until we are sure that there is no retraction before taking the action, then in what sense are we doing early classification — we are surely just doing classification.

We believe that the prefix problem may be essentially insurmountable in many domains. For example, imagine we wanted to early classify the *vocalization* of {gun, point}. There are eighty-eight English words beginning with gun, including *gunwales*, *gunnel*, *gunnysack*, *gunk*, etc., and twenty-six words that begin with point, including *pointedly*, *pointlessness*, *pointier*, *pointman*, etc.

### 3.2 The Inclusion Issue

**The inclusion problem** is the assumption that the pattern to be early classified is not comprised of smaller atomic units that are frequently observed on their own.

For example, suppose we learn a model for early classification of the vocalization of {lightweight, paperweight}. We can do very well after seeing the first 10% to 20% of these utterances (which is fortunate, as the final 54% of the signal is identical and offers no additional information). However, suppose the universe contains sentences such as "*In the morning light, I could see that I got a papercut from the paper that the light was wrapped in.*" This sentence would give us two false positives for each class. Moreover, it is clear that the sub-pattern could be vastly more common than the full modeled pattern. For words, this is simply an obvious implication of Zipf's law.

Returning to our vocalization of {gun, point} example, recall that in English, we will encounter words like *disappointing*, *ballpoints*, *appointment*, *burgundy*, *begun*, etc., and also proper names like *Gunderson*, the *Pointer* sisters, etc.

### 3.3 The Homophone Issue

**The homophone problem** is the assumption that two semantically different events will have different shapes in the time series representation.

Suppose that we learn a model for early classification of the vocalization of {flower, wither}. Moreover, we are fortunate that in this problem space, we are told that *any* word containing the target word is also a true positive, so we should take the same action for *flower*, *flowerpot*, *deflowered*, and for *wither*, *witheringly, swithering*, etc. This means we are completely free of the prefix and inclusion problems above. However, what are we to make of the following sentence from Leviticus 2:1 "*Whither anyone presents a grain offering as an offering to the Lord, his offering shall be of fine flour, and…*"? This sentence does not contain either of the target words, but it contains two near-perfect homophones, *flower* vs. *flour* and *wither* vs. *whither*, which would give us false positives.

Just because we know that the semantic meaning of the classes in which we are interested is different, it does not follow that the time series representation we see will also be different. For example, as shown in Fig. 6 and Fig. 9, *gun* and *point* are extracted from video by tracking the center of mass of the right hand. They are sufficiently different to be distinguished with high accuracy. However, it is possible that completely different behaviors such as *removing-spectacles*, *looking-at-watch*, or *lighting-a-cigarette* are perfect "homophones" in the time series space. In fact, given the vast space of human actions, the very limited one-dimensional view of 150 datapoints virtually assures us this will be the case.

In order to show that *time series* homophones exist, we conducted the following experiment. We randomly selected two examples from the GunPoint dataset, and for each of them, we searched for its three nearest neighbors. However, rather than searching within a human behavior dataset, we searched within three datasets that do *not* have gestures. Fig. 5 shows the results.

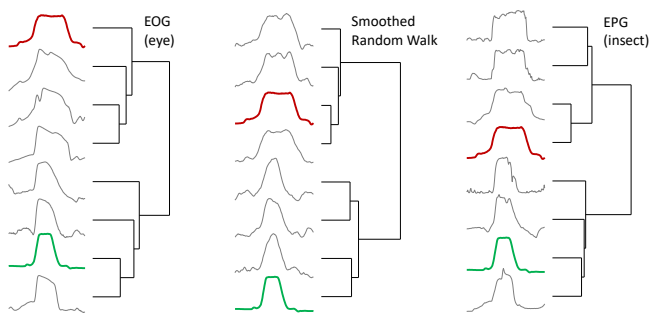

Fig. 5. Two random examples from the GunPoint dataset (colored), clustered with their nearest neighbors from: *(left)* One hour of eye movement data; *(center)* A smoothed random walk of length $2^{24}$; *(right)* Eight hours of insect behavior.

Note that in every case, there is non-gesture data that is much closer to one member of the target class, than the *other* example from the target class. We can repeat this experiment with all datasets from the UCR archive with similar results.

The homophone problem can also show up as part of the inclusion problem. For example, the last author occa-sionally searches Google for *shapelets*, the time series primitive. Most of the hits are true positives, but Google also returns pages with "*Unique puzzle piece shape lets it interlock with…*", "*A simple shape lets the beauty of the faux concrete…*", and "*Its triangular shape lets you reach the corners of the pool…*". So even though the word *shapelets* does not have a homophone, it does have pseudo-homophones[1]. If we simply searched a large text corpus, we would surely find a lot more of these pseudo-homophones than hits to the obscure data mining primitive.

Returning to our vocalization of {gun, point} example, recall that in English, we will encounter words like *pointe*, *pint*, *Gunn* (proper name), etc.

### 3.4 Summary for this Section

We believe that the prefix, inclusion, and homophone problems imply the space of possible domains where ETSC could be meaningfully applied is vanishingly small. Again, returning to the problem of the vocalization of {gun, point} for a final time. A single English sentence such as "*Amy Gunn thought it pointless to go on pointe before she had begun her appointment to get her burgundy ballet shoes cleaned off all the gunk…*" would produce a plethora of false positives. While most of our examples are contrived for ease of exposition, Fig. 5 suggests these problems are common in real-valued time series, as does a more general exploration of the datasets in the UCR Archive [1].

It is important to note that while our examples used natural language for simplicity, we have observed these issues in datasets containing gestures, writing, electrical power demand, chicken behavior, insect behavior, bird vocalizations, and in almost everywhere we looked.

There is a data domain *that* might be free of these issues: electrocardiograms (ECGs), photoplethysmograms, and similar time series. However, in the next section, we will show that all ETSC papers that report apparently good results on these datasets are inadvertently "cheating" by peeking into the future.

## 4 PEEKING INTO THE FUTURE

Almost all papers on ETSC suffer from a logical flaw that means that their accuracy would plunge if we attempted to use them on streaming data[2]. Once again, the UCR format is the culprit. The UCR datasets are z-normalized. However, when you see the prefix of an oncoming pattern in a streaming environment, you cannot z-normalize it until *after* you have seen all the data, which of course, means that you are not doing *early* classification.

Many researchers seem unaware of just how brittle distance measures are to changes in the mean (and standard deviation) of the exemplars. To show this, let us revisit GunPoint. As shown in Fig. 6, we produced a "denormalized" version of the test data by adding to each instance a random number in the range [-1, 1].

---

[1] Yes, multiple pseudo-homophones: *Our plush ape lets you dress him*.
[2] Paper [2] does *not* have this flaw. The current authors warned them of this issue before [2] was published.

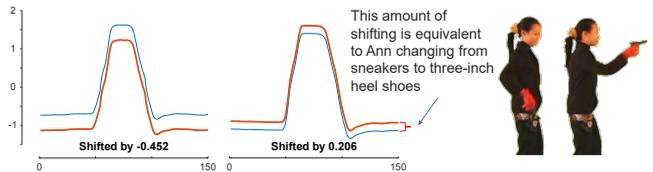

Fig. 6. Original examples from the GunPoint dataset together with denormalized versions, which have been slightly shifted in the Y-axis.

It is important to understand how small of a change this is. It is approximately equivalent to tilting the camera randomly up or down by about 1.9 degrees. Or it is equivalent to replacing Ann with Jessica, a slighter taller grad student.

It is also important to note what effect this would have on normal nearest neighbor classification: *none*. It has long been known that you should z-normalize the data before computing the Euclidean distance or DTW [24]. In Table 1, we compute the accuracy of six ETSC algorithms on the UCR-normalized data and the denormalized data. We used the authors' recommended settings and/or tested many settings and reported only the *best* results.

TABLE 1
The accuracy of six early classification algorithms

| Algorithm | Normalized | DeNormalized |
|---|---|---|
| (*min. support* = 0) ECTS [3] | 86.7% | 68.7% |
| (*min. support* = 0) RelaxedECTS [3] | 86.7% | 68.7% |
| EDSC-CHE [4] | 94.7% | 62.7% |
| EDSC-KDE [4] | 95.3% | 58.7% |
| ($\tau$ = 0.1) Rel. Class. [8] | 90.0% | 70.0% |
| ($\tau$ = 0.1) LDG Rel. Class. [8] | 91.3% | 71.3% |

These results show that the algorithms can do apparently very well on GunPoint. However, when we apply the model to streaming data, if the camera zooms in or out, or tilts up or down, or one of the actors decides to go barefoot, or the actor stands a little closer to the camera, etc., the accuracy will plunge.

It is critical not to misunderstand this result. It is not that these algorithms forgot a step, and we can just add it back in. When the algorithms see a value, they are assuming that it is z-normalized based on other values that do not yet exist! As we noted above, ECGs are a favorite example for ETSC papers [2], [3], [4], [5], [6], [7], [8], [9], [10], [11], [12], [13], [14], [15], [16], [17], [18], [19], [20]. In Fig. 7, we show a tiny snippet of an ECG (recorded from two different chest locations) before it was contrived into the UCR data format.

The practical upshot of this problem is that these algorithms working on medical telemetry will be plagued with false negatives. One might try to get past this issue by saying, "*well, the models will work for domains that don't need z-normalization*". However, Rakthanmanon *et al.* [24] make a forceful case that such domains are very rare or nonexistent.

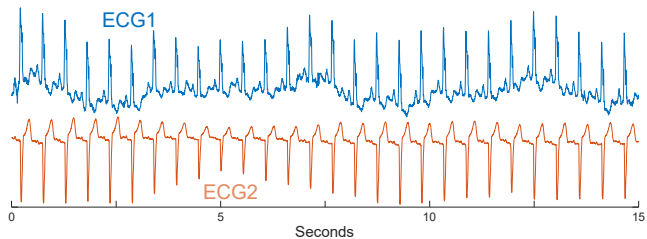

Fig. 7. An ECG recorded from two locations in the chest. ECG1 shows dramatic but medically meaningless variation in the mean of individual beats. ECG2 shows equally dramatic but also medically meaningless variation in the standard deviation of individual beats.

## 5 Does Early Classification *Ever* Make Sense?

In our long search for a dataset that might work under ETSC assumptions, our best match was a dataset that consists of more than 12.5 billion datapoints of chicken behavior, measured using a "backpack" accelerometer, as shown in Fig. 8 *(right)*.

Consider the time series shown in Fig. 8 *(left)*. It is an excellent template to detect the behavior of dustbathing in chickens. Any subsequence that is within 2.3 of z-normalized Euclidean distance of this template is essentially guaranteed to be dustbathing.

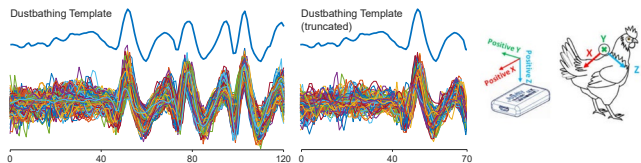

Fig. 8. *(left)* A template for dustbathing and its 500 nearest neighbors. *(center)* A truncated version of the template and its 500 nearest neighbors. *(right)* The data was obtained from a backpack sensor.

The time series shown in Fig. 8 *(center)* is a prefix of the first template, and any subsequence that is within 1.7 of this template can be classified as dustbathing with an accuracy that is not statistically significantly different from the accuracy achieved with the longer template.

One can even make a case for actionability here. Suppose you want to prevent the chicken from conducting long periods of dustbathing. Perhaps if you early classify a dustbathing behavior, you could flash a bright light, or play the sound of a chicken's alarm cackle, either one of which would startle the chicken out of its intended behavior. Note that the cost of a false positive is not too high here (although it is not zero, chickens do become desensitized to frequent alarms).

Have we found an example that justifies ETSC? Perhaps, but consider:

- A reader might reasonably say that this is not *early* classification, but rather simply classification with an awareness of the obvious fact that the sensitivity and specificity of time series template will (typically non-linearly) change as you add or delete points to either end.
- We did not need any special algorithms or models to understand that the shorter template is as effective as the longer template. This took common sense and a few minutes of low-code exploration of the data.
- No data from this domain was ever placed into the UCR format. At a minimum, discovering template(s) would need to be done *before* we could even attempt to put the data into the UCR format.

Clearly, absence of evidence is not evidence of absence. But it is surprising that it is so difficult to find a dataset where ETSC would make sense. More telling, to the best of our knowledge, no one in the community has produced a publicly available dataset where it can be claimed: ETSC would be useful, and some ETSC models have been shown to work.

Finally, at the risk of appearing cynical, it is easy to see that one could use this dataset to write a paper that apparently shows utility for ETSC. We could massage more examples like the longer template in Fig. 8 *(left)* into the UCR format and show our "model" learns to predict dustbathing after seeing only 70% of the data! Such a claim would look very impressive, but it is only with the context above that we realize that the claim would be vacuous. Could similar situations explain other apparent ETSC successes?

With this in mind, let us revisit the GunPoint dataset, which is particularly beloved by the ETSC community [2], [3], [4], [6], [7], [10], [11], [12], [13], [15], [18], [20]. We have deep insights into this dataset, because the third author created it in 2003. In order to create a simple-to-use dataset, we used a metronome to synchronize the performance of the behaviors (pointing or aiming). The metronome sounded a "beep" every five seconds, and the two "actors" were given the following brief: "*When you hear the cue, wait about a second, do the behavior for about two seconds, then return your hand to the side for the remaining time.*" As shown in Fig. 9, this means that the last one to two seconds of most of the GunPoint exemplars are non-informative and non-class discriminating sections where the hand was resting by the actors' side. In addition, as hinted by the dataset's name, the difference between the classes is mostly the actors' fumbling to remove the gun from the hostler, which happens at the *beginning* of the action.

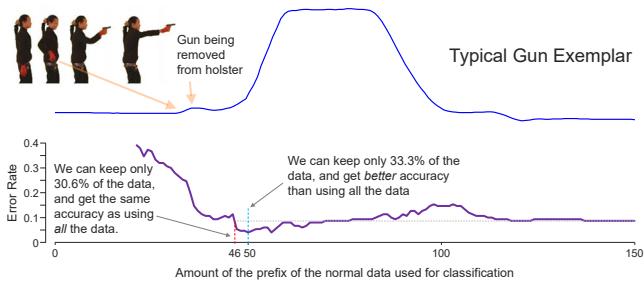

Fig. 9. *(top)* A typical example from GunPoint annotated to show where the discriminating region is. *(bottom)* The holdout classification error-rate of every prefix of the GunPoint data from lengths 20 to 150 (the full length of the data).

The plot shown in Fig. 9 *(bottom)* resembles many plots shown in ETSC papers (actually, it is better than most of them, as we are correctly z-normalizing the truncated data, see Table 1). However, it is important to note that we are not claiming contribution by this plot, this is just basic data cleaning, not a publishable research model.

Note that a large number of UCR datasets have similar formatting conventions, some "events" bookended by constant regions that are simply there to make all the data objects have the same length (CricketX, CBF, Trace, etc.). Thus, it seems possible that some (possibly a very large) fraction of the apparent success of ETSC may be due to nothing more than a formatting convention that padded the right side of events with uninformative data, just to make the objects the same length.

## 6 CONCLUSIONS

The time series early classification task as commonly understood may not be a meaningful problem to solve. All current research efforts that address this problem will be condemned to being overwhelmed by false positives if actually deployed in a real-world setting. Of course, false positives are a fact of life for any machine learning problem. However, the unique claims of *immediate actionability* mean that these false positives will have a cost, and the false positives may be many orders of magnitude more common than true positives. In addition, virtually all the algorithms are making the assumption that the data they are seeing *now* is normalized relative to data that only exists in the *future*. All those algorithms are condemned to producing mostly false negatives.

We believe that the issue is not with the proposed algorithms per se. The issue is that the definition of the problem itself is intrinsically underspecified and vague. The following are our recommendations to bring clarity to the ETSC area:

- An effort should be made to provide a concrete, testable, falsifiable, and useful definition of early classification of time series. While the current authors have no interest in providing this definition (in any case, a consortium of researchers would be better), we believe that any such definition would, at a minimum, have to consider:
  1) The cost of a false positive for the actionable class(es) vs. the cost of a false negative [12], [19]. Even if the only early action taken is to sound an alarm, false alarm fatigue is known to have a high cost [21].
  2) The probability that the domain of interest contains *prefixes*, *inclusions*, and *homophones* that resemble the actionable class(es).
  3) The prior probability of seeing a member of the actionable class(es).
  4) The appropriateness of the normalization assumptions for the domain.
- Anyone proposing an ETSC model needs to carefully explain what the model offers beyond simply classification with trivial awareness that not all datapoints matter (recall Fig. 9).
- It is hard to see how any genuine progress could be made without access to a real-world publicly available dataset(s) that could benefit from the more concrete definition. The overreliance on the UCR datasets seems to have led the community astray here. Proxy datasets and synthetic datasets *do* have their place in research, especially in fledging areas. However, we are now two decades and many dozens of papers into this area.

It is hard to overemphasize the last point. If no real-world publicly available dataset(s) where some form of ETSC is useful can be obtained, this seems tantamount to saying that there is no problem to solve, and the community should stop publishing on this topic. It is stunning to think of the ease with which a grad student can obtain seismic data recorded on Mars, or the mitochondria DNA of a mammoth that has been extinct for a million years, yet everyone publishing on ETSC must resort to proxy datasets.

## APPENDIX A

## ON THE TERM *EARLY CLASSIFICATION*

The term "Early Classification" is unfortunately overloaded and vague. There are several tasks that might be named as such, which do not fall under the purview of this paper. For example:

- Suppose that a boiler is rated for at most 200 psi. If a sensor detects increasing pressure readings: 180, 181, 182, ..., it would make perfect sense to sound an early warning that the pressure may approach 200 psi. Note that this setting only considers the *value* of a time series, not the *shape* of the time series. The same is true for many medical domains: if a person's BMI is measured monthly and begins to creep up to 20, 21, 22, ..., it might be better for a doctor to suggest an intervention before it reaches 25. But again, only the *value*, not the *shape* matters.
- Monitoring of batch processes is a slight generalization of the above. At every time point in a single run (plus or minus some "wiggle room" that can be modeled [25]), we know what range of values are acceptable. If the reading begins to drift outside that range, we can sound an alarm. Once again, this problem only considers the *value* of a time series, not the *shape* of the time series.
- Suppose that a chicken engaging in dustbathing more than 40 times a day is required to be culled by local ordinance (because dustbathing is often caused by the presence of mites or other pests) [26]. If we detect 10 bouts of dustbathing one day and 25 the next day, we may want to take some early intervention. Note that this setting only considers the *frequency* of (fully observed, not "early" observed) behaviors.

More generally, there may be other problems that have been labeled "early classification" by someone. We make no claims about such work. Our claims are limited to the sense of early classification used in [2], [3], [4], [5], [6], [7], [8], [9], [10], [11], [12], [13], [14], [15], [16], [17], [18], [19], [20], where the prefix of the *shape* of the time series is assumed to contain information that we can act upon before seeing the remainder of the shape.

# APPENDIX B
## OBJECTIONS TO OUR CLAIMS

Given the unusual nature of our claims, we solicited feedback from the community while writing this paper. We did this by writing to every author that published a paper on ETSC, and by general postings on discussion boards such as *www.reddit.com/r/MachineLearning/*

Most of the feedback has been (gratefully) incorporated into the main text. Here, we respond to a few questions that are worth addressing but would spoil the flow of our paper:

- **Q)** *Doesn't the fact that there are commercial predictive text algorithms for <u>handwriting</u> tell us that the prefix/ inclusion/homophone problems can be overcome?*
  **A)** These systems are not doing predictive classification based on *words*, they are classifying individual *letters*, and then using classic ASCII predictive text algorithms. Moreover, as the Google help page notes *"Stand-alone symbols that are just a line (1/l/I) or circle (o/O/0) can be difficult to distinguish"* [27], exactly because those groups of symbols appear as homophones in time series space.

- **Q)** *Your claim "it is not clear any of them could ever work in a real-world setting" seems too strong.*
  **A)** Let us clarify what it means for a model to "work" here. Simply producing plots like Fig. 8 is not sufficient. Every event we are trying to detect has a cost. For concreteness, let us consider petrochemical engineering, and say the target event is the undesirable foaming of a distillation column. Assume it costs $1,000 to clean out the apparatus after such an event. Let us further imagine that if we get "early" notice that this is about to happen, we can warn an engineer to throttle some valve, and stop the damage. This action must also have some cost, let us say $200. Thus, in order for an ETSC model to be said to work, it must at least break even, producing at least one true positive for every five false positives. A handful of ETSC papers do have costs built into their models [12], [19], but they only test on UCR datasets and never estimate costs for any real-world applications. The results shown in this paper suggest that the vast majority of positives will be false positives. For example, we applied the model in [2] to the Gun-Point problem, with the exemplars inserted in between long stretches of random walks, and we see thousands of false positives for every true positive (see [28]).

- **Q)** *Doesn't the homophone problem imply that all time series classification is hard, not just ETSC?*
  **A)** Yes, it does to some extent. Even if you ignore the issues of *early* classification, and consider only *classic* time series classification, the UCR datasets seem to have led to an illusion of progress. However, at least some applications do bypass this problem. For example, there are many papers on using the time series obtained from the sensors in a Wii Remote to classify gestures as inputs to the system. Normally, the user presses a button that indicates "start classifying" and releases it once the gesture is recognized. This means that the algorithm is not asked to deal with spurious data that might be thousands of times more frequent than target data. Such uses of time series classification *do* largely fit into the UCR format assumptions. Likewise, objects that come from the *spectrogram* and (converted from 2D) *shape* datatypes are presented as discrete vectors, not part of a stream.

- **Q)** *I don't see why z-normalization would be imperative in all real problems.*
  **A)** We think this question has been addressed in [24] and elsewhere by the community. However, in brief: it *is* meaningful to compare time series based on z-normalized shape; it is *sometimes* meaningful to compare time series based on mean value; but it is almost never meaningful to cluster on *both* at the same time (which is equivalent to comparing non-normalized time series with shape measures). The reason is that even small differences in the mean (and/or the standard deviation) completely drown out any shape information. In other words, for *non-normalized* data, $dist(\text{mean}(a), \text{mean}(b)) \propto dist(a, b)$, where *dist* is the Euclidean distance or DTW, etc. To summarize, if z-normalization is not important in your domain, it is virtually certain that the *shapes* do not matter — only the absolute values do. We make no claim about such situations other than the obvious empirical observation that such domains are very rare.

## ACKNOWLEDGMENT

The authors wish to thank the many individuals who gave us feedback. However, any errors remaining in this paper are ours alone. All datasets and code used in this paper are archived in perpetuity at [28].

**Renjie Wu** is currently a PhD candidate in Computer Science at the University of California, Riverside. He received his B.S. degree in Computer Science and Technology from Harbin Institute of Technology at Weihai in 2017. His research interests include time series analysis, data mining, and machine learning.

**Audrey Der** is currently a PhD student in Computer Science at the University of California, Riverside, from which she received her B.S. degree in Computer Science in 2019. Her research interests include time series analysis, data mining, and machine learning.

**Eamonn Keogh** is a professor of Computer Science at the University of California, Riverside. His research interests include time series data mining and computational entomology.